\documentclass[conference]{IEEEtran}
\IEEEoverridecommandlockouts
\usepackage{cite}
\usepackage{amsmath,amssymb,amsfonts}
\usepackage{algorithm, algorithmic}
\usepackage{graphicx}
\usepackage{subcaption}
\usepackage{multirow}
\usepackage{textcomp}
\usepackage{xcolor}
\usepackage{hyperref}
\def\BibTeX{{\rm B\kern-.05em{\sc i\kern-.025em b}\kern-.08em
    T\kern-.1667em\lower.7ex\hbox{E}\kern-.125emX}}
    
\begin{document}
\title{DINOSTAR: Deep Iterative Neural Object Detector Self-Supervised Training for Roadside LiDAR Applications\\
}
\author{\IEEEauthorblockN{Muhammad Shahbaz}
\IEEEauthorblockA{\textit{CECE} \\
\textit{University of Central Florida}\\
Orlando, Florida \\
Muhammad.Shahbaz@ucf.edu}
\and
\IEEEauthorblockN{Shaurya Agarwal}
\IEEEauthorblockA{\textit{CECE} \\
\textit{University of Central Florida}\\
Orlando, Florida \\
shaurya.agarwal@ucf.edu}
\and
\IEEEauthorblockN{Karan Sikka}
\IEEEauthorblockA{
\textit{SRI International}\\
Princeton, USA \\
karan.sikka@sri.com}
}

\maketitle

\begin{abstract}
Recent advancements in deep-learning methods for object detection in point-cloud data have enabled numerous roadside applications, fostering improvements in transportation safety and management. However, the intricate nature of point-cloud data poses significant challenges for human-supervised labeling, resulting in substantial expenditures of time and capital. This paper addresses the issue by developing an end-to-end, scalable, and self-supervised framework for training deep object detectors tailored for roadside point-cloud data. The proposed framework leverages self-supervised, statistically modeled teachers to train off-the-shelf deep object detectors, thus circumventing the need for human supervision. The teacher models follow fine-tuned set standard practices of background filtering, object clustering, bounding-box fitting, and classification to generate noisy labels. It is presented that by training the student model over the combined noisy annotations from multitude of teachers enhances its capacity to discern background/foreground more effectively and forces it to learn diverse point-cloud-representations for object categories of interest. The evaluations, involving publicly available roadside datasets and state-of-art deep object detectors, demonstrate that the proposed framework achieves comparable performance to deep object detectors trained on human-annotated labels, despite not utilizing such human-annotations in its training process.
\end{abstract}

\begin{IEEEkeywords}
self-supervised, object detection, roadside, lidar, point-cloud data, student-teacher modeling
\end{IEEEkeywords}

\section{Introduction}
\begin{figure}[h!t!]
\centering
\centerline{\includegraphics[width=\linewidth]{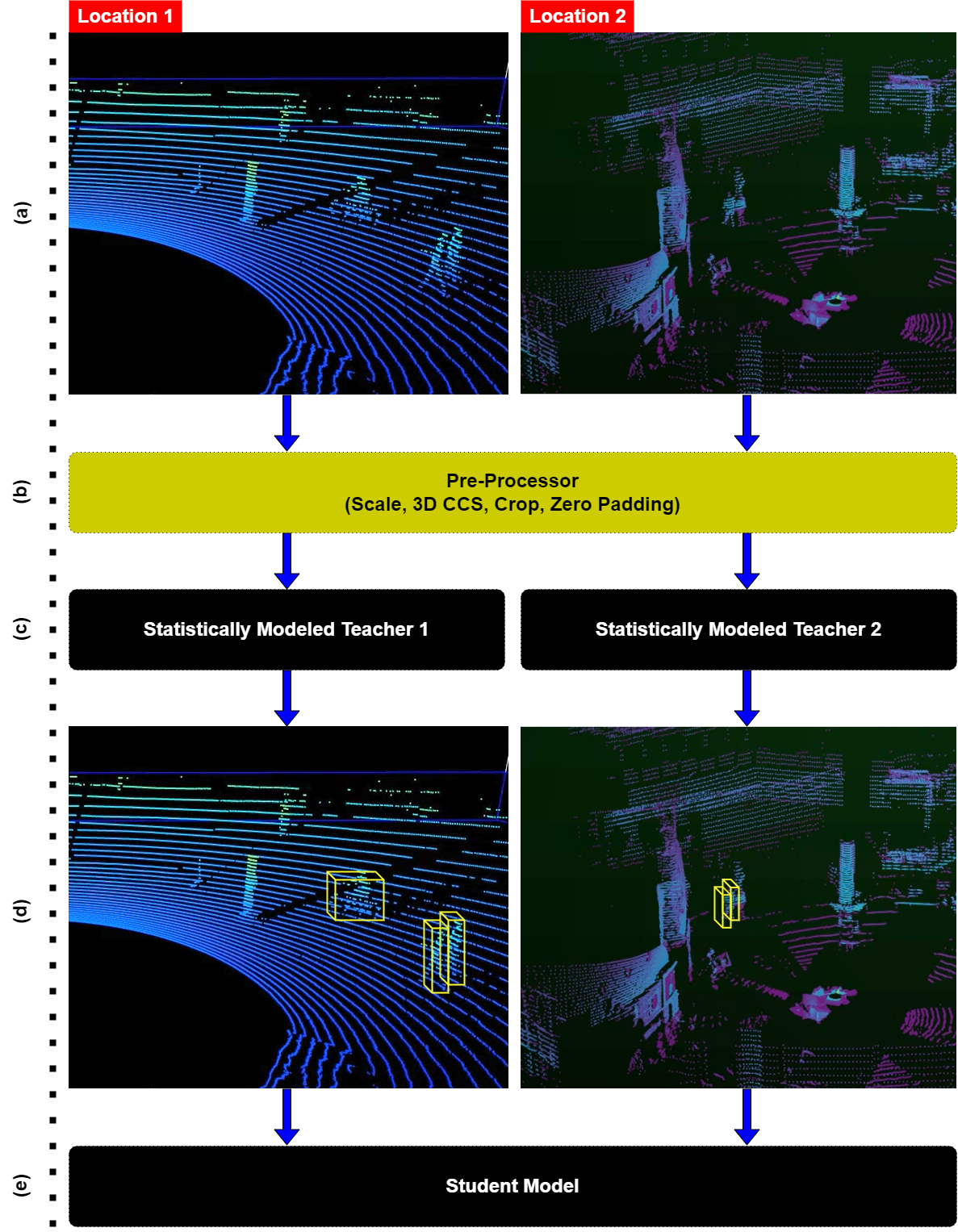}}
\caption{We gather point cloud data from multiple location using similar sensors (a), the data is unified in pre-processing stage (b) and use separate (by changing hyper-parameters of) statistically modeled teachers (c) to generate a superset of weakly-labeled datasets (d). We then train the deep object detector (student model) to generalize over the superset (e).}
\label{fig:generic-object-detectors}
\end{figure}
In recent literature, the task of object detection in pure point cloud spaces, that is, detecting and classifying objects of interest by regressing a bounding-box around the object and assigning a category label, is dominantly solved by deep neural network models \cite{guo2020deep}. Such models are data hungry \cite{chen2014big}, \cite{torralba2011unbiased} and, often, require human-annotated labels during training process to achieve satisfactory performance \cite{sun2017revisiting}. Consequently, many human-annotated point cloud datasets have been made available over the years. In autonomous vehicle domain, datasets such as KITTI \cite{geiger2013vision}, nuscenes \cite{caesar2020nuscenes}, Waymo Open \cite{sun2020scalability}, Pandaset \cite{xiao2021pandaset}, and Apollo \cite{bakogiannis2019apollo} stand as prominent examples. However, the data distributions in those datasets do not conform to stationary roadside LiDARs because of the differences in perspective (position and looking angle) and focus (regions of interest). As a result, models trained on vehicle-mounted lidar datasets can not be directly used for roadside lidar data \cite{wu2018automatic}, \cite{wang20213d}. Therefore, there has been a notable surge in human-annotated datasets focusing roadside LiDAR applications recently, including, IPS300+ \cite{wang2022ips300+}, BAAI-VANJEE \cite{yongqiang2021baai}, Rope3D \cite{ye2022rope3d}, Dair-v2x \cite{yu2022dair}, LUMPI \cite{busch2022lumpi}, and A9 \cite{cress2022a9}. However, the publication of such datasets incur substantial expenditures both in terms of time and money due to higher data dimentionality of point-clouds. This, quickly, becomes infeasible as the roadside LiDAR infrastructure scales up, introducing many perspectives and foci depending on sensor placement. Therefore, a more automated and cost-effective approach to train deep object detectors for roadside LiDARs is required.

\begin{figure*}
\centering
\includegraphics[width=\linewidth]{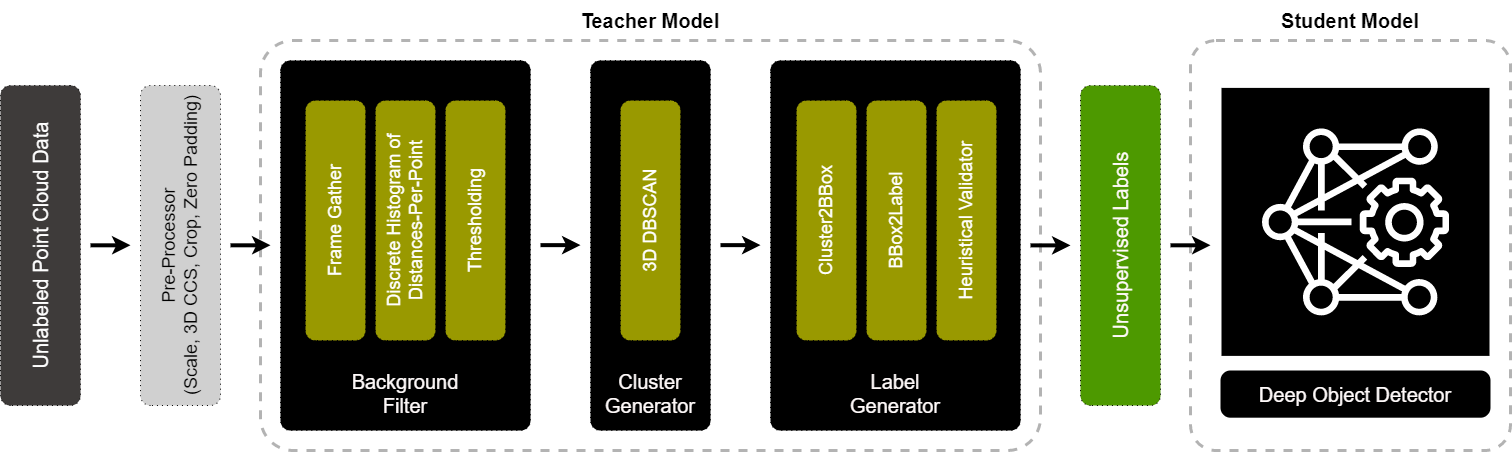}
\caption{Label Generation Pipeline: The teacher model (left) uses a novel background filtering method followed by traditional DBSCAN clustering to generate object clusters that are then classified based on shape heuristics to create annotations that are then used by student model (right) as ground-truth labels.}
\label{fig:all-stgs}
\end{figure*}

In this paper, we present a scalable, end-to-end, self-supervised framework for training object detectors for roadside stationary LiDAR sensors. The framework utilizes teacher-student modeling approach to train a more robust deep object detector (student model) utilizing the labels generated by a set of self-supervised yet noisy statistically modeled detectors (teacher models). The framework is illustrated in figure \ref{fig:all-stgs}. First, (unannotated) point-cloud datasets of a geographic location are captured from multiple perspectives and foci. For each dataset, a separate teacher model is employed to generate noisy labels. Here, the teacher uses standard yet heuristically fine-tuned standard practices of automated object detection in point-clouds including background filtering, object clustering, bounding-box fitting, and heuristics-based classification to generate noisy labels. The generated labels are then combined and are utilized as ground-truth annotations by a student model (a deep object detector) during its training process. Here, we propose that such modeling approach benefits from (1) more data, as the bottleneck introduced by human-supervised labeling is removed, and (2) more data diversification, as data annotation can be parallelized for multiple locations, perspectives, and foci by using multitude of teacher models. This can be thought of as leveraging ensemble of teacher models to train a highly resilient student model by capitalizing on their collective knowledge. The entire code of the project is available at our \textbf{[Link to our Github Repo]}. It is important to note that although the proposed framework idealizes multitude of teachers processing data of the same geographical location from multiple perspective and foci, it does not necessitate the use of multiple teachers or the same geographical location. Moreover, the framework is also agnostic to the choice of deep object detector (student model) however application specific deep object detectors demonstrated better performance. This has been studied extensively in appendix I of the paper. 

The contribution of this paper are as following:
\begin{enumerate}
    \item It presents a novel self-supervised framework to train deep object detectors for roadside point-cloud data.
    \item It develops tool chain to train point-cloud deep object detectors in an end-to-end fashion.
    \item It develops a novel strategy to automate object detection for roadside applications at large scales.
    \item It is first to evaluates the results on IPS300+\cite{wang2022ips300+}, the largest roadside dataset available to date.
\end{enumerate} 

The remainder of this manuscript is arranged in the following manner: Section II presents an overview of the existing literature relevant to our study. This is followed by an explanation of our research methodology in Section III. The experimental procedures and their subsequent results are explored in Section IV. In the final segment, Section V, we draw conclusions from our research and outline prospective areas for further exploration.

\section{Related Work}
Until very recently, the research on roadside lidar object detection remained hindered due to lack of common benchmark for performance evaluation. However, since 2021, the roadside lidar research has got the spot light among computer vision researchers as more and more roadside lidar datasets are getting published \cite{brucker2022unsupervised}. Currently, most of the literature in domain focus on either traditional machine learning methods or deep learning based object detection methods \cite{sun2022object}.

\textbf{Traditional Methods}: The traditional methods utilizing machine learning usually involve four steps: 1) background filtering, 2) foreground object point clustering, 3) feature extraction, and 4) classification. Each of these steps is crucial but background filtering and foreground object point clustering, perhaps, are the most important steps that affects the overall accuracy of the model. The background filtering step, usually, differs based on the input point cloud representation (range image, voxels, azimuth-height table, dynamic-matrix, adaptive-grid, elevation azimuth matrix, etc.). The foreground object point clustering differs based on clustering method used (euclidean clustering, distance-based, DBSCAN, and modified versions of DBSCAN). Sun et. al. \cite{sun2022object} discussed it all in details in their comprehensive review. 

\textbf{Deep Learning Methods:}
In recent years, various methods have been proposed for object detection in roadside LiDAR data based on deep learning. Wang et al. \cite{wang20213d} utilized a background filtering module and 3D CNN detectors, with a strategy of filtering out background points before model training to enhance the generalization ability and performance of 3D detectors. Zhang et al. \cite{zhang2022} employed background filtering, clustering, a tracker module, and a PointVoxel-RCNN detector, generating candidate objects via background filtering, moving point clustering, and a UKF tracker before detection. Zhou et al. \cite{zhou2022} modified the PointPillars architecture, reusing CNNs pre-trained on autonomous driving data for vehicle detection from roadside LiDAR data and introducing dense connections between convolutional layers to achieve more effective features. Bai et al. \cite{bai2022} used a combination of RPEaD, PointPillars, and FPN, transforming roadside point clouds into coordinates suitable for training on the onboard dataset, voxelizing the point cloud to generate point cloud pillars, and designing a feature pyramid network for bounding box prediction. Zimmer et al. \cite{zimmer2022} introduced DASE-ProPillars, an improved version of the PointPillars model, with strategies including introducing extensions to improve PointPillars, creating a semi-synthetic roadside LiDAR dataset for model training, manually labeling LiDAR frames for detector fine-tuning, and performing transfer learning.

\textbf{Hybrid Methods:}
Some works have begun to combine traditional methods with deep learning models for object detection in roadside LiDAR data. For instance, Wang et al. \cite{wang20213d} proposed a method that involves filtering out the background points before training a 3D Convolutional Neural Network (CNN) detector. Similarly, Zhang et al. \cite{zhang2022optimizing} combined background filtering, clustering, and a tracker module with a PointVoxel-RCNN detector for object detection. These methods demonstrate the potential of integrating traditional and deep learning techniques for improved object detection performance. However, to the best of our knowledge no prior studies combine objection detection methods from traditional machine learning and deep learning object detectors in teacher-student modeling approach to achieve completely self-supervised training.

In the next section, we will delve into the methodology of our study, building upon the insights gained from these related works.

\section{Methodology}
The teacher model in the proposed framework has four sub-modules: (1) pre-processor, (2) background filter, (3) clustering algorithm, and (4) bounding-box regressor and object classifier. The student model can be any pure point-cloud-based deep object detectors. Currently, we are using OpenPCDet \textbf{[refer to OpenPCDet paper]} implementation of SECOND[reference] as our deep object detector. The former is discussed in detail in the following, the later, however, is only discussed shortly at the end of this section as it has insignificant contributions to the proposed methodology.
\subsection{The Teacher Model}
The input data to the teacher model is a spatio-temporal set $X = \{\mathbf{x}_1,\mathbf{x}_2,\mathbf{x}_3,\dots,\mathbf{x}_T\}$ of point cloud frames where each frame is a set $\mathbf{x}_t = \{p_1, p_2, p_3, \dots, p_N\}$ of 3D points at time $t \in [1, 2, 3, \dots, T]$.

\subsubsection{Pre-Processor}
Since the unit of measurements and point cloud representations may differ across different LiDAR sensors, the spatio-temporal set $X$ of point cloud frames is first unified by scaling points per metric system and representation per 3D Cartesian Coordinate System (CCS), followed by zero padding each frame such that $|x_t|$ for $t \in [1, 2, 3, \dots, T]$ is a fixed number, represented by a hyper-parameter $N_\text{total}$. Then, each frame $\mathbf{x} \in X$ is cropped spatially to form $X^\text{cropped} = \{\bar{\mathbf{x}}_1,\bar{\mathbf{x}}_2,\bar{\mathbf{x}}_3,\dots,\bar{\mathbf{x}}_T\}$ such that:

\begin{equation}
\mathbf{\bar{x}} = q(\mathbf{x}), \label{eq_X_query_ar}
\end{equation}
\[
\text{where},\;q(\mathbf{x}) = 
\begin{cases}
x_{\min} \leq p_x \leq x_{\max} \\
y_{\min} \leq p_y \leq y_{\max} \\
z_{\min} \leq p_z \leq z_{\max} \\
\end{cases}
\forall \; p \in \mathbf{x}
\]

Here, $q(\mathbf{x})$ represents a cuboid bound parameterized by the minimum $(x_{\min}, y_{\min}, z_{\min})$ and maximum $(x_{\max}, y_{\max}, z_{\max})$ values of its three dimensions. Thus $X^\text{cropped}$ contains only those points from set $X$ that fall within the specified 3D range defined by $q$.

\subsubsection{Background Filter}
In the case of non-translating LiDAR sensor, the background and foreground points can be segmented rather easily as the background regions are non-moving with respect to the LiDAR sensor. It is exploited in background filter module of teacher model to generate a set of background-removed point cloud frames $X^\text{foreground}$. In the proposed method it is done by, first, extracting a set of query frames of size $N_\text{query}$ by:

\begin{equation}
X^\text{query} = X^\text{cropped}[1,2,3 \dots, N_\text{query}], 
\label{eq:x_query}
\end{equation}

Then, a Discrete Histogram of Distances-Per-Point (D-HistDPP), $H^\text{distance}$, parameterized by $N_\text{bin}$ number of bins, is calculated using algorithm \ref{alg:discrete-hist-of-dist-per-point}. $H^\text{distance}$ represents the point cloud as a density matrix of per-point distances over time constrained by $N_\text{query}$. Hence, the value of $N_\text{query}$ is directly related to quality of background/foreground estimation, however, it leads to more computation cost. Similarly, increasing number of bins, that is $N_\text{bin}$, decreases discretization leading to a granular background/foreground estimation. However, too much discretization results in lesser fault tolerance that happens due to small jitters in sensor.

\begin{algorithm}
\caption{D-HistDPP: Discrete Histogram of Distances-Per-Point}
\label{alg:discrete-hist-of-dist-per-point}
\begin{algorithmic}[1]
\REQUIRE $(X^\text{query}, N_\text{query}, N_\text{total}, N_\text{bin})$ \\
\ENSURE $H^\text{distance}$
\STATE $P^\text{distance} \gets \textbf{zeros}[N_\text{query}, N_\text{total}]$
\FOR{$1 \leq i \leq N_\text{query}$}
    \FOR{$1 \leq j \leq N_\text{total}$}
        \STATE $P^\text{distance}_{i,j} \gets \|X^\text{query}_{i,j}\|$
    \ENDFOR
\ENDFOR
\STATE $P^\text{distance} \gets P^\text{distance}.\textbf{T}$
\STATE $H^\text{distance} \gets \boldsymbol{\phi}[N_\text{total}, N_\text{bin}]$

\FOR{$1 \leq i \leq N_\text{total}$}
    \STATE $d_\text{max} \gets \textbf{max}(P^\text{distance}_i)$
    \STATE $d_\text{min} \gets \textbf{min}(P^\text{distance}_i)$
    \STATE $w \gets \frac{d_\text{max} - d_\text{min}}{N_\text{bin}}$
    \FOR{$1 \leq j \leq N_\text{query}$}
        \STATE $k \gets \lfloor \frac{P^\text{distance}_{i,j} - d_\text{min}}{w} \rfloor$
        \STATE $H^\text{distance}_{i,k} \gets H^\text{distance}_{i,k} + P^\text{distance}_{i,j}$
    \ENDFOR
    \FOR{$1 \leq k \leq N_\text{bin}$}
        \STATE $H^\text{distance}_{i,k} \gets \textbf{mean}(H^\text{distance}_{i,k})$
    \ENDFOR
\ENDFOR
\end{algorithmic}
\end{algorithm}

The background filtering algorithm \ref{alg:background-filtering} uses $H^\text{distance}$ to construct $X^\text{foreground}$. First, $H^\text{distance}$ is sorted in descending order of heights (number of distances in each bin). Then, $N_\text{tall}$ number of tallest bins are extracted. Finally, $X^\text{foreground}$ is constructed by removing every point $p \in \bar{x} \in X^\text{cropped}$ if the difference between its distance and its corresponding distance in $H^\text{distance}$ is less than a hyper-parameter $D_\text{threshold}$.

\begin{algorithm}
\caption{Background Filtering}
\label{alg:background-filtering}
\begin{algorithmic}[1]
\REQUIRE $(X^\text{cropped}, H^\text{distance}, N_\text{total}, N_\text{tall}, D_\text{threshold})$ \\
\ENSURE $X^\text{foreground}$
\STATE $H^\text{distance} \gets \textbf{sort}(H^\text{distance})$
\FOR{$1 \leq i \leq N_\text{total}$}
\STATE $H^\text{distance}_i \gets H^\text{distance}_{i}[1, N_\text{tall}]$
\ENDFOR
\STATE $X^\text{foreground} \gets \boldsymbol{\phi}$
\FORALL{$\bar{\mathbf{x}} \in X^\text{cropped}$}
    \STATE $\hat{\mathbf{x}} \gets \phi$
    \FOR {$1 \leq i \leq N_\text{total}$}
        \STATE $\bar{d} \gets \|\bar{\mathbf{x}}_i\|$
        \FORALL{$\tilde{d} \in H^\text{distance}$}
            \IF{$|\bar{d} - \tilde{d}| \leq D_\text{threshold}$}
                \STATE $\hat{\mathbf{x}} \gets \hat{\mathbf{x}} + \bar{\mathbf{x}}_i$
                \STATE \textbf{break}
            \ENDIF
        \ENDFOR
    \ENDFOR
    \STATE $X^\text{foreground} \gets X^\text{foreground} + (\bar{\mathbf{x}} - \hat{\mathbf{x}})$
\ENDFOR
\end{algorithmic}
\end{algorithm}

Intuitively, $N_\text{tall}$  represents the number of background location one point-ray can hit; a higher value allows better mapping of background locations in case there is foliage in the environment. Moreover, the threshold $D_\text{threshold}$ act as radius of a region around a background location center that is considered part of the that background object, allowing the algorithm to handle small noise in retrieved point locations from background objects.

\subsubsection{Clustering Algorithm}
The clustering module of the teacher model takes $X^\text{foreground}$ as input to create per-frame-object clusters, represented by a set $Y^\text{clusters}$, that can be further processed for object detection and classification. Ideally, $X^\text{foreground}$ contains points reflected from object-of-interest (Vehicles, Pedestrians, Cyclists) in roadside scene, however it is not true. Due to small movements in background objects of the scene (for example, trees) and the inherent noise of the sensor, a significant number of noisy points still exist in the $X^\text{foreground}$. To handle clustering in presence of noisy points, a 3D version of DBSCAN (Density-Based Spatial Clustering of Applications with Noise) algorithm \cite{ester1996density} is utilized. DBSCAN is a density-based clustering method. It takes three main parameters: the point set $\bar{\mathbf{x}} \in X^\text{foreground}$, the distance threshold $\varepsilon$, and the minimum number of points $MinPts$.

\subsubsection{Object Classifier \& BBox Regressor}
The object clusters generated as result of DBSCAN algorithm, $Y^\text{clusters}$, are of various sizes and shapes. Here, the teacher model encapsulates each cluster with an minimal-bounding box and takes three heuristical hyper-parameters: 1) bbox minimum base-length $l_\text{min}$, 2) bbox minimum height $h_\text{min}$, and 3) minimum difference between bbox base-length and height $\beta_\text{min}$ to validate the bbox. The validated bbox are then classified based on a very simple heuristic, that is, vehicle's base-length is greater than its height whereas for pedestrians it is vice versa.

\subsection{Student Model}
The student model employed in this study refers to a uni-modal deep object detector that specifically operates on point cloud data. For our case studies, we utilized the OpenPCDet implementation of Sparsely Embedded Convolutional Detection (SECOND) as one of our chosen models [reference]. Additionally, other models were also considered [mention additional models].

\begin{figure}[bh]
\centerline{\includegraphics[width=\linewidth]{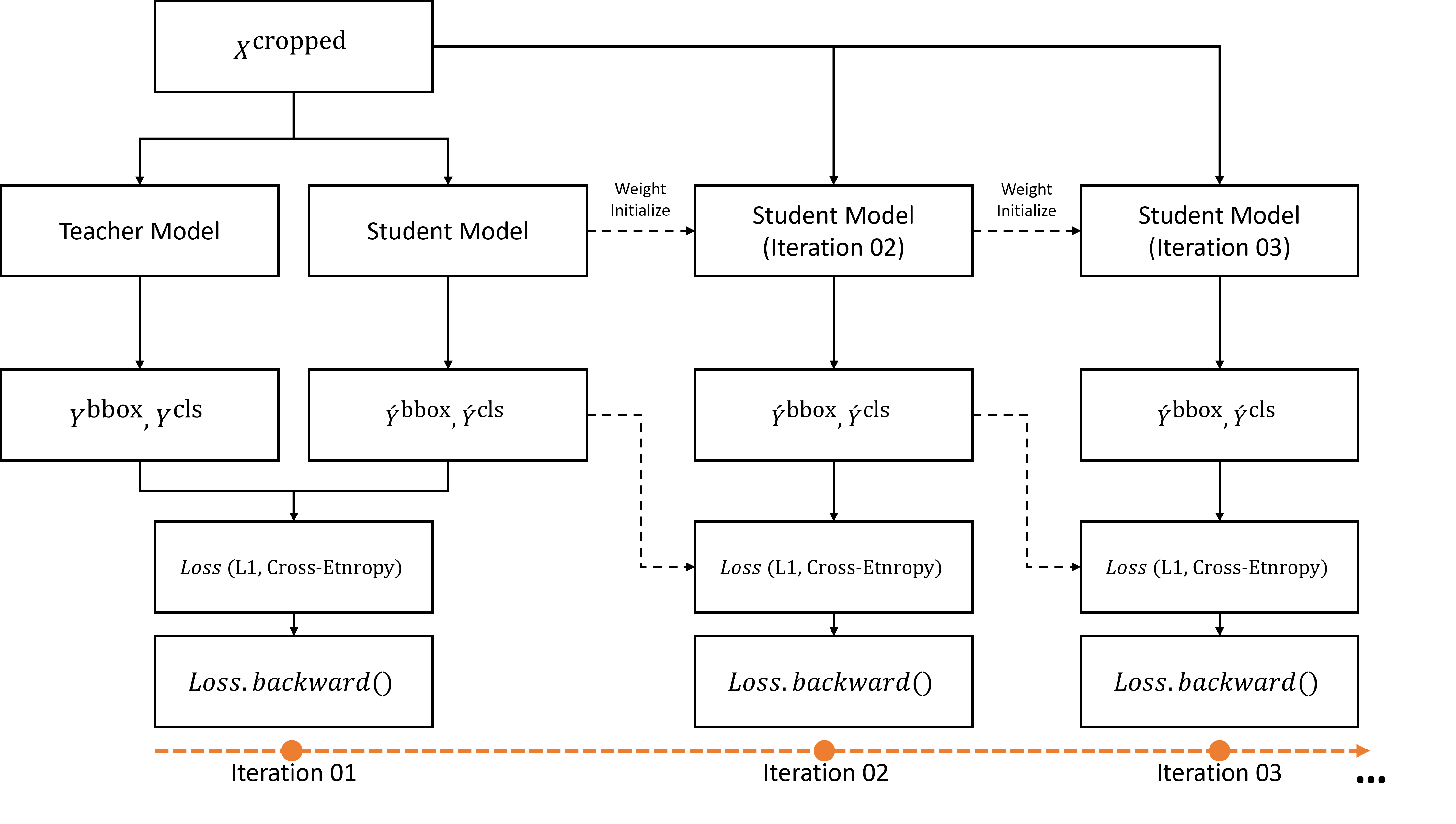}}
\caption{Iterative Training for Improving Student Belief}
\label{fig:iterative-learning}
\end{figure}

The input to the deep detector consists of cropped point cloud frames denoted as $X^\text{cropped}$, with the objective of predicting the bounding box labels $Y^\text{bbox}$ and the classification labels $Y^\text{cls}$ for two classes: 1) Vehicles and 2) Pedestrians. Following training, the student model can be further enhanced through an iterative learning process. This iterative learning process involves leveraging the trained model to generate ground-truth labels, which are then used for training in subsequent iterations, as depicted in Figure \ref{fig:iterative-learning}.

\section{Experiments \& Results}
\textbf{Baseline Datasets: }For baseline benchmarking, we employed four datasets: 1) Pedsafe (ours), 2) ASU Pedestrian LiDAR Scenes (APLS) dataset, 3) Providentia++ A9, and 4) IPS300+.
These datasets focus small intersection, inside campus, freeway traffic, and large intersection respectively and are recorded using three LiDAR sensors with different specifications \ref{tbl:data-specs}.

\begin{table}[htbp]
  \centering
  \caption{Data Specifications}
  \label{tbl:data-specs}
  \begin{tabular}{|c|c|c|c|c|c|}
    \hline
    \multicolumn{2}{|c|}{\textbf{Dataset}} & \textbf{Pedsafe} & \textbf{APLS} & \textbf{A9} & \textbf{IPS300+} \\
    \hline
    \multicolumn{2}{|c|}{\textbf{Category}} & Intersection & Campus & Freeway & Intersection \\
    \hline
    \multicolumn{2}{|c|}{\textbf{Height}} & 10ft & 6ft? & 22.96ft & 18ft \\
    \hline
    \multicolumn{2}{|c|}{\multirow{2}{*}{\textbf{Sensor Name}}} & Ouster & Velodyne & Ouster & Robosense\\
                     \multicolumn{2}{|c|}{} & OS1-64 & HDL-32E & OS1-64 & Ruby-Lite\\
    \hline
    \multirow{2}{*}{\textbf{VFOV}} & Lower & -22.5°& -20.665° & -16.55° & -25° \\
    \cline{2-6}
    & Upper & +22.5° & +20.665° & +16.55° & 15° \\
    \hline
    \multicolumn{2}{|c|}{\textbf{Resolution}} & 1024x64 & ~1000x32 & 1024x64 & 360x80 \\
    \hline
    \multicolumn{2}{|c|}{\textbf{Frequency}} & 10Hz & 10Hz? & 10Hz & 5Hz \\
    \hline
  \end{tabular}
\end{table}

Such diversity in data poses a challenge and is therefore selected for case studies of our framework. However, we limit our target classes to 1) vehicles and 2) pedestrians for the sake of simplicity of training procedure. Therefore, any extra labels in both datasets are removed.

For a through evaluation, three training and testing experiments were conducted. First, we used ~9000 unlabeled frames from Pedsafe dataset, passed it through teacher model to generate weak labels and trained student model (SECOND) and performed evaluations on ~1000 human-labeled test frames from Pedsafe dataset. 

\textbf{Supersets: }Subsequently, we added more data to the initial train dataset by auto-annotating roughly 40,000 frames from APLS dataset (also unlabeled) by another teacher model. However, combining multiple datasets has its challenges. Therefore, first both the datasets were unified by centering both at rougly the same point and scaling so object in both dataset are of (roughly) same sizes; at our Github \textbf{Github Link}, we provide the tools for such unification operations. The combined train dataset is used for training a second student model (also SECOND) and evaluated on ~1000 human-labeled test frames from Pedsafe dataset.

The results present a huge improvement of roughly 20 scores on mAP@0.25 metric for pedestrian detection task. Finally, we utilized the model trained in last attempt (Pedsafe + APLS) and tested it against A9 and IPS300+ dataset. However, due to huge difference in perspectives the model performance dropped drastically. The results are presented in table \ref{tab:evaluation-scores}.

\begin{table}[htbp]
  \centering
  \caption{Dataset Splits}
  \label{tbl:dataset-splits}
  \begin{tabular}{|c|c|c|c|}
    \hline
    \textbf{Dataset} & \textbf{Used in Training} & \textbf{Used in Testing} & \textbf{Split} \\
    \hline
    Pedsafe & \checkmark & $\times$ & 8,690 \\
    \hline
    Pedsafe & $\times$ & \checkmark & 966 \\
    \hline
    \multirow{1}{*}{APLS} & \checkmark & $\times$ & 39,917 \\
    \hline
    \multirow{1}{*}{A9} & $\times$ & \checkmark & 456 \\
    \hline
    \multirow{1}{*}{IPS300+} & $\times$ & \checkmark & 1,000 \\
    \hline
  \end{tabular}
\end{table}

\begin{table}[htbp]
  \centering
  \caption{Evaluation Scores for Pedestrian Detection}
  \label{tab:evaluation-scores}
  \begin{tabular}{|c|c|c|c|c|c|}
    \hline
    \multicolumn{2}{|c|}{\textbf{Training}} & \textbf{mAP} & \textbf{mAP} & \textbf{Recall} & \textbf{Recall} \\
    \multicolumn{2}{|c|}{\textbf{Dataset}} & \textbf{@0.5} & \textbf{@0.25} & \textbf{@0.5} & \textbf{@0.3} \\
    \hline
    \multirow{2}{*}{Pedsafe} & Pedsafe & 15.293 & 40.7668 & 25.9784 & 79.8981 \\
    \cline{2-6}
    & Pedsafe + APLS & 30.165 & 61.3668 & 34.8043 & 82.9309 \\
    \hline
    \multirow{2}{*}{A9} & Pedsafe & 0 & 0 & 0 & 0.861 \\
    \cline{2-6}
    & Pedsafe + APLS & 0 & 0 & 0 & 0.861 \\
    \hline
    \multirow{2}{*}{IPS300+} & Pedsafe & 0.3588 & 6.2761 & 0.062525 & 0.273771 \\
    \cline{2-6}
    & Pedsafe + APLS & 0.9091 & 3.6576 & 0.029092 & 0.113172 \\
    \hline
  \end{tabular}
\end{table}

\subsection{Iterative Training}
The teacher model utilizes statistics and heuristics for detecting and labeling objects in the scene limiting the student model to only perceive weak labels. It is proposed that since student model learns the combined distribution from ensemble of teacher models (in case of combining multiple datasets experiments), the generalized student model has better understanding of background removal than individual teacher models. Here, we propose that the if student model is forced to increase its own belief, by increasing the probability of predicting the labels generated by itself, the student become more resilient to the errors in individual teacher models. This can be achieved by simply generating labels using student model (i.e., inferencing student model on input point cloud frames), and then using those generated labels as ground-truth labels to train a successive student model (a student model initialized on weights and biases of previous student); the concept is illusterated in figure \ref{fig:iterative-learning}.

\section{Conclusion and Future Work}

In this work, we presented a scalable, end-to-end, self-supervised framework for training object detectors for roadside stationary LiDAR sensors. The framework utilizes a teacher-student modeling approach to train a more robust deep detector (student model) based on the labels generated by a self-supervised yet noisy statistically modeled detector (teacher model). The proposed method benefits from more data, as the bottleneck introduced by human-supervised labeling is removed, and more data diversification, as data annotation can be parallelized for multiple locations, perspectives, and foci by using a multitude of teacher models. 

The iterative process initially improves the accuracy of the student model. However, we observed that after three iterations, the accuracy no longer exhibits significant improvement. This study provides all the necessary tools to convert, unify, and auto annotate (teacher model) pure point-cloud datasets, and to train and evaluate deep object detector (student model).

In future work, we aim to further improve the accuracy of the student model beyond the observed three iterations. We also plan to explore the potential of integrating additional traditional and deep learning techniques for improved object detection performance. Furthermore, we intend to investigate the feasibility of applying the proposed methodology to other types of sensors and detection tasks.

\bibliography{main} 

\begin{thebibliography}{10}

\bibitem{bai2022}
Z.~Bai, S.P. Nayak, X.~Zhao, G.~Wu, M.J. Barth, X.~Qi, and K.~Oguchi.
\newblock Cyber mobility mirror: Deep learning-based real-time 3d object
  perception and reconstruction using roadside lidar.
\newblock {\em arXiv preprint arXiv:2202.13505}, 2022.

\bibitem{bakogiannis2019apollo}
Tasos Bakogiannis, Ioannis Giannakopoulos, Dimitrios Tsoumakos, and Nectarios
  Koziris.
\newblock Apollo: A dataset profiling and operator modeling system.
\newblock In {\em Proceedings of the 2019 International Conference on
  Management of Data}, pages 1869--1872, 2019.

\bibitem{brucker2022unsupervised}
Marcel Brucker.
\newblock Unsupervised lidar-based 3d object detection using infrastructure
  sensors, 2022.

\bibitem{busch2022lumpi}
Steffen Busch, Christian Koetsier, Jeldrik Axmann, and Claus Brenner.
\newblock Lumpi: The leibniz university multi-perspective intersection dataset.
\newblock In {\em 2022 IEEE Intelligent Vehicles Symposium (IV)}, pages
  1127--1134. IEEE, 2022.

\bibitem{caesar2020nuscenes}
Holger Caesar, Varun Bankiti, Alex~H Lang, Sourabh Vora, Venice~Erin Liong,
  Qiang Xu, Anush Krishnan, Yu~Pan, Giancarlo Baldan, and Oscar Beijbom.
\newblock nuscenes: A multimodal dataset for autonomous driving.
\newblock In {\em Proceedings of the IEEE/CVF conference on computer vision and
  pattern recognition}, pages 11621--11631, 2020.

\bibitem{chen2014big}
Xue-Wen Chen and Xiaotong Lin.
\newblock Big data deep learning: challenges and perspectives.
\newblock {\em IEEE access}, 2:514--525, 2014.

\bibitem{cress2022a9}
Christian Cre{\ss}, Walter Zimmer, Leah Strand, Maximilian Fortkord, Siyi Dai,
  Venkatnarayanan Lakshminarasimhan, and Alois Knoll.
\newblock A9-dataset: Multi-sensor infrastructure-based dataset for mobility
  research.
\newblock In {\em 2022 IEEE Intelligent Vehicles Symposium (IV)}, pages
  965--970. IEEE, 2022.

\bibitem{ester1996density}
Martin Ester, Hans-Peter Kriegel, J{\"o}rg Sander, Xiaowei Xu, et~al.
\newblock A density-based algorithm for discovering clusters in large spatial
  databases with noise.
\newblock In {\em kdd}, volume~96, pages 226--231, 1996.

\bibitem{geiger2013vision}
Andreas Geiger, Philip Lenz, Christoph Stiller, and Raquel Urtasun.
\newblock Vision meets robotics: The kitti dataset.
\newblock {\em The International Journal of Robotics Research},
  32(11):1231--1237, 2013.

\bibitem{guo2020deep}
Yulan Guo, Hanyun Wang, Qingyong Hu, Hao Liu, Li~Liu, and Mohammed Bennamoun.
\newblock Deep learning for 3d point clouds: A survey.
\newblock {\em IEEE transactions on pattern analysis and machine intelligence},
  43(12):4338--4364, 2020.

\bibitem{sun2017revisiting}
Chen Sun, Abhinav Shrivastava, Saurabh Singh, and Abhinav Gupta.
\newblock Revisiting unreasonable effectiveness of data in deep learning era.
\newblock In {\em Proceedings of the IEEE international conference on computer
  vision}, pages 843--852, 2017.

\bibitem{sun2020scalability}
Pei Sun, Henrik Kretzschmar, Xerxes Dotiwalla, Aurelien Chouard, Vijaysai
  Patnaik, Paul Tsui, James Guo, Yin Zhou, Yuning Chai, Benjamin Caine, et~al.
\newblock Scalability in perception for autonomous driving: Waymo open dataset.
\newblock In {\em Proceedings of the IEEE/CVF conference on computer vision and
  pattern recognition}, pages 2446--2454, 2020.

\bibitem{sun2022object}
Pengpeng Sun, Chenghao Sun, Runmin Wang, and Xiangmo Zhao.
\newblock Object detection based on roadside lidar for cooperative driving
  automation: a review.
\newblock {\em Sensors}, 22(23):9316, 2022.

\bibitem{torralba2011unbiased}
Antonio Torralba and Alexei~A Efros.
\newblock Unbiased look at dataset bias.
\newblock In {\em CVPR 2011}, pages 1521--1528. IEEE, 2011.

\bibitem{wang20213d}
G.~Wang, J.~Wu, T.~Xu, and B.~Tian.
\newblock 3d vehicle detection with rsu lidar for autonomous mine.
\newblock {\em IEEE Transactions on Vehicular Technology}, 70:344--355, 2021.

\bibitem{wang2022ips300+}
Huanan Wang, Xinyu Zhang, Zhiwei Li, Jun Li, Kun Wang, Zhu Lei, and Ren
  Haibing.
\newblock Ips300+: a challenging multi-modal data sets for intersection
  perception system.
\newblock In {\em 2022 International Conference on Robotics and Automation
  (ICRA)}, pages 2539--2545. IEEE, 2022.

\bibitem{wu2018automatic}
Jianqing Wu, Hao Xu, Yuan Sun, Jianying Zheng, and Rui Yue.
\newblock Automatic background filtering method for roadside lidar data.
\newblock {\em Transportation Research Record}, 2672(45):106--114, 2018.

\bibitem{xiao2021pandaset}
Pengchuan Xiao, Zhenlei Shao, Steven Hao, Zishuo Zhang, Xiaolin Chai, Judy
  Jiao, Zesong Li, Jian Wu, Kai Sun, Kun Jiang, et~al.
\newblock Pandaset: Advanced sensor suite dataset for autonomous driving.
\newblock In {\em 2021 IEEE International Intelligent Transportation Systems
  Conference (ITSC)}, pages 3095--3101. IEEE, 2021.

\bibitem{ye2022rope3d}
Xiaoqing Ye, Mao Shu, Hanyu Li, Yifeng Shi, Yingying Li, Guangjie Wang, Xiao
  Tan, and Errui Ding.
\newblock Rope3d: the roadside perception dataset for autonomous driving and
  monocular 3d object detection task.
\newblock In {\em Proceedings of the IEEE/CVF Conference on Computer Vision and
  Pattern Recognition}, pages 21341--21350, 2022.

\bibitem{yongqiang2021baai}
Deng Yongqiang, Wang Dengjiang, Cao Gang, Ma~Bing, Guan Xijia, Wang Yajun, Liu
  Jianchao, Fang Yanming, and Li~Juanjuan.
\newblock Baai-vanjee roadside dataset: Towards the connected automated vehicle
  highway technologies in challenging environments of china.
\newblock {\em arXiv preprint arXiv:2105.14370}, 2021.

\bibitem{yu2022dair}
Haibao Yu, Yizhen Luo, Mao Shu, Yiyi Huo, Zebang Yang, Yifeng Shi, Zhenglong
  Guo, Hanyu Li, Xing Hu, Jirui Yuan, et~al.
\newblock Dair-v2x: A large-scale dataset for vehicle-infrastructure
  cooperative 3d object detection.
\newblock In {\em Proceedings of the IEEE/CVF Conference on Computer Vision and
  Pattern Recognition}, pages 21361--21370, 2022.

\bibitem{zhang2022optimizing}
J.~Zhang, W.~Xiao, and J.P. Mills.
\newblock Optimizing moving object trajectories from roadside lidar data by
  joint detection and tracking.
\newblock {\em Remote Sensing}, 14:2124, 2022.

\bibitem{zhang2022}
S.~Zhou, H.~Xu, G.~Zhang, T.~Ma, and Y.~Yang.
\newblock Leveraging deep convolutional neural networks pre-trained on
  autonomous driving data for vehicle detection from roadside lidar data.
\newblock {\em IEEE Trans. Intell. Transp. Syst.}, 23:22367--22377, 2022.

\bibitem{zhou2022}
S.~Zhou, H.~Xu, G.~Zhang, T.~Ma, and Y.~Yang.
\newblock Leveraging deep convolutional neural networks pre-trained on
  autonomous driving data for vehicle detection from roadside lidar data.
\newblock In {\em Proceedings of the IEEE/CVF Conference on Computer Vision and
  Pattern Recognition}, pages 10529--10538, 2022.

\bibitem{zimmer2022}
W.~Zimmer, M.~Grabler, and A.~Knoll.
\newblock Real-time and robust 3d object detection within road-side lidars
  using domain adaptation.
\newblock {\em arXiv preprint arXiv:2204.00132}, 2022.

\end{thebibliography}
\bibliographystyle{plain}

\end{document}